\documentclass[11pt]{article}

\usepackage[final]{acl}

\usepackage{times}
\usepackage{latexsym}
\usepackage{booktabs}
\usepackage{multirow}
\usepackage{amsmath}
\usepackage[T1]{fontenc}
\usepackage{subcaption}
\usepackage{placeins}
\usepackage[utf8]{inputenc}

\usepackage{microtype}

\usepackage{inconsolata}

\usepackage{graphicx}
\usepackage[table,dvipsnames]{xcolor}

\newcommand{\best}[1]{\cellcolor{Gray!35}\textbf{#1}}
\newcommand{\second}[1]{\cellcolor{Gray!25}#1}
\newcommand{\third}[1]{\cellcolor{Gray!10}#1}
\newcommand{\fourth}[1]{\cellcolor{Gray!5}#1}

\title{Cross-lingual Functional Vectors for Emotion Detection \\in Large Language Models}

  \author{
  Jieying Xue \and
  Phuong Minh Nguyen \and
  Minh Le Nguyen \and
 Shogo Okada 
\\
  Japan Advanced Institute of Science and Technology 
\\
\texttt{\{xuejieying,phuongnm,nguyenml,okada-s\}@jaist.ac.jp}
\\
}

\begin{document}
\maketitle
\begin{abstract}
Function vectors (FVs) have recently emerged as a promising mechanism for steering the behavior of large language models (LLMs) by injecting task-specific latent direction representations derived from in-context demonstrations.
While prior studies have shown that FVs can recover task behavior in structured in-context learning settings, their effectiveness on semantically complex tasks and their ability to generalize across languages remain underexplored.
We investigate the \textit{cross-lingual transferability of FVs} using multilingual multi-label emotion recognition as a challenging semantic classification benchmark.
Specifically, we examine whether FVs extracted from a source language can steer task behavior in another language under both standard clean and perturbed zero-shot settings without providing demonstrations during inference.
Across diverse cross-lingual settings, applying FVs substantially improves performance, suggesting that FVs capture language-agnostic, task-relevant signals rather than purely language-specific lexical patterns, and highlighting their potential as a lightweight and transferable mechanism for multilingual task adaptation.
We observe that each LLM exhibits a relatively stable optimal range of attention heads for constructing effective FVs, and the pattern remains consistent across languages.
In addition, FVs can partially replicate the task-steering effects of standard few-shot in-context learning while avoiding the computational overhead of processing multiple demonstrations, making them effective for large-scale practical applications. Our code is available at \url{https://github.com/yingjie7/cross_lingual_fvs}.

\end{abstract}

\section{Introduction}
Large language models (LLMs) exhibit remarkable in-context learning (ICL) abilities, enabling them to perform new tasks from only a few demonstrations without parameter updates \cite{brown2020language,adcock2026llama,yang2025qwen3,singh2025openai}.
Recent studies suggest that this capability is partially encoded in latent activation patterns within the LLMs.
In particular, function vectors (FVs) have been proposed as latent task representations that steer model behavior through activation interventions at intermediate layers of LLMs \cite{todd2024function}.
Existing studies have demonstrated the effectiveness of FVs on relatively constrained tasks, such as simple classification, retrieval, and factual induction \cite{todd2024function}. 
However, whether FVs remain effective for semantically complex tasks that require richer contextual understanding and reasoning remains largely unexplored. 
Therefore, this work investigates FVs using multilingual multi-label emotion recognition as a challenging semantic classification benchmark. Unlike the relatively simple tasks considered in previous work, multilingual emotion recognition requires 
implicit intention understanding, and the disambiguation of subtle affective expressions across substantial linguistic variation \cite{muhammad-etal-2025-semeval,Belay2024EvaluatingTC}. This task provides a challenging testbed for evaluating whether FVs encode transferable task-level representations beyond relatively simple pattern matching.

Beyond task complexity, we further investigate the cross-lingual transferability of FVs. Prior studies have shown that multilingual LLMs exhibit partially shared language-agnostic latent representations across languages \cite{zhang-etal-2024-unveiling-linguistic,zeng-etal-2025-converging,bandarkar2025layer}, while other work suggests that these representations may remain biased toward English-centered latent structures \cite{zhao2024how,wendler2024llamas,schut2025multilingual,tezuka2025transfer}.
These observations naturally raise an important question: \textit{Do FVs encode language-specific behaviors, or transferable task-level representations that generalize across languages?} If the latter is true, FVs extracted from one language should be able to recover and transfer task behavior to other languages without requiring target-language demonstrations during inference.

To answer this question, we conduct a comprehensive study of FVs under both \textit{zero-shot} and \textit{few-shot} prompting settings.
In the zero-shot setup, we evaluate FVs under both standard clean prompts and a perturbed zero-shot setting. The perturbed setting introduces controlled distractor labels into the prompt, allowing us to isolate the task-steering effect of FVs and evaluate whether cross-lingual FVs can recover the intended task behavior despite misleading contextual information. Experimental results demonstrate that FVs extracted from one language can be effectively transferred to multiple target languages, and are able to steer task behavior under both clean and perturbed zero-shot conditions. 
We further investigate whether FVs provide complementary task representations under multilingual few-shot prompting. Specifically, FVs are injected into the induction heads of LLMs during inference and shared across languages. While FVs do not completely replace the benefits of standard few-shot ICL, they consistently enhance multilingual few-shot performance and partially reproduce the task-steering effects without requiring additional demonstrations during inference. Moreover, we observe that each LLM exhibits a relatively stable range of attention heads for constructing effective FVs, and that this pattern remains consistent across languages.
Overall, our findings suggest that FVs encode transferable task-level representations that are less dependent on language-specific lexical patterns, highlighting their potential as an efficient and transferable mechanism for multilingual task adaptation.
Our contributions are summarized as follows: 
\begin{itemize}
\item We extend FV evaluation from relatively constrained tasks to a semantically complex multilingual setting. We show that FVs remain effective on tasks requiring contextual semantic reasoning, suggesting that the limitations observed in prior work are not inherent to FVs themselves.
\item We provide a systematic analysis of FV intervention across transformer layers and show that distributing FV interventions across multiple layers substantially outperforms single-layer intervention, offering a practical strategy for applying FVs to complex semantic tasks.
\item We present the first systematic study of cross-lingual FVs for multilingual semantic classification, demonstrating effective task transfer under both standard clean and controlled perturbed zero-shot prompting.
\item We analyze the stability of attention-head selection for FVs across languages and find that each LLM exhibits a stable range of attention heads for constructing effective FVs, and that this pattern remains consistent across languages.
\item We demonstrate that cross-lingual FVs complement multilingual few-shot prompting. Although they do not fully replace standard few-shot ICL, cross-lingual FVs consistently improve multilingual few-shot performance and partially reproduce the task-steering effects of in-context learning without requiring additional demonstrations during inference.
\end{itemize}

\section{Related Work}

\subsection{FVs and Mechanistic Interpretability}

Recent studies on ICL suggest that task behavior in LLMs may be partially encoded in internal activation patterns rather than solely induced by surface-level prompts \cite{todd2024function,hendel-etal-2023-context,wang2023label, guo2024transformers,Liu2023IncontextVM}. 
In particular, prior work interprets ICL as constructing latent task representations from demonstrations \cite{hendel-etal-2023-context}, while other studies analyze how task-relevant information propagates through models from an information-flow perspective \cite{wang2023label}.
Within the growing field of mechanistic interpretability, FVs have been proposed as compact latent representations that capture task-inducing signals in LLMs \cite{todd2024function}.
These studies demonstrate that injecting FVs into intermediate layers can steer model behavior toward task-consistent outputs, even under zero-shot settings, and that only a small subset of attention heads plays a critical role in transporting task-relevant information.
Related approaches, such as in-context vectors (ICVs) \cite{Liu2023IncontextVM}, similarly manipulate hidden representations extracted from demonstrations to guide model behavior.
However, existing studies primarily evaluate FVs and related steering methods on relatively structured monolingual tasks. Prior work reported limited effectiveness of FVs on semantically complex tasks, such as sentiment analysis and synonym prediction, where task boundaries are less structurally explicit \cite{todd2024function}. Consequently, whether FVs remain effective for multilingual semantic classification tasks requiring richer contextual reasoning and cross-lingual semantic understanding remains largely unexplored.

\subsection{Cross-lingual Representation in LLMs}

Understanding how multilingual LLMs represent and transfer knowledge across languages has become an important research direction.
Prior work suggests that multilingual LLMs learn partially shared latent representations across languages while preserving language-specific characteristics \cite{qin2025survey,zhang-etal-2024-unveiling-linguistic,zeng-etal-2025-converging,bandarkar2025layer}. Other studies further show that multilingual reasoning may rely on shared latent representations, whereas language-specific generation behavior tends to emerge in later transformer layers \cite{zhao2024how,wendler2024llamas,schut2025multilingual,tezuka2025transfer}.
From a mechanistic perspective, recent studies have begun to analyze multilingual capabilities at the level of internal activations. 
For example, Cross-Lingual Activation Steering (CLAS) demonstrates that selectively modulating neuron activations can improve multilingual performance without updating model parameters \cite{pokharel2026cross}.
Likewise, prior work shows that both language-specific and language-general attention heads contribute to multilingual processing, suggesting that multilingual behavior is partially localized within model internals \cite{liu2026focusing}.
Despite these advances, existing studies mainly investigate language-level representations or multilingual reasoning as a whole. Whether task-level latent representations themselves can be transferred across languages remains insufficiently explored. In contrast, our work investigates whether FV can recover and transfer task behavior across languages in semantically complex multilingual classification settings.

\section{Method}
We investigate whether FVs encode language-independent task representations that generalize across multilingual emotion classification settings.
To this end, we evaluate cross-lingual transferability of FVs under both \textit{zero-shot} and \textit{few-shot} settings.
Specifically, we examine whether FVs extracted from a source language can recover task behavior in target languages without target-language demonstrations under both standard clean and the perturbed zero-shot setting, and whether they provide complementary task representations when combined with standard few-shot ICL.

\paragraph{Problem Formalization.} 
We define a dataset in a language $g$: $\mathcal{D}_{g} = \{(x_{i g}, y_{i g})\}_{0\leq i < |\mathcal{D}_{g}|}$, where $g \in \mathcal{G} = \{ en, ch, ...\}$.
Given the input prompt $p$, an LLM ($f$) is expected to decode the label $y$.
Under the ICL setting, the cross-lingual few-shot prompt is formulated as an \textit{informative} (clean) prompt:
\begin{equation} 
\small
p^{\mathrm{few}_\mathrm{g_1 g_2}} = \left[ (x_{1 \mathrm{g_1}}, y_{1 \mathrm{g_1}})... (x_{N \mathrm{g_1}}, y_{N \mathrm{g_1}}), x_{q\mathrm{g_2}} \right]
\end{equation}
where $g_1$ and $g_2$ denote the source language of the demonstrations and the target language of the query input, respectively. In the remainder of this paper, unless otherwise specified for $g_1$ and $g_2$, the demonstrations and query share the same language; that is, $g_1 = g_2$. Note that in zero-shot prompting, no demonstrations are provided; therefore, only $g_2$ is used. In contrast, during FV extraction, only the language used in the demonstrations, $g_1$, is utilized. 

We design two experimental settings to examine the cross-lingual capabilities of FVs:
(1) comparing \textit{Zero-shot} prompting in language $g_2$ with \textit{Zero-shot + FVs$^{g_1}$} to evaluate the ability of FVs to encode task directions and steer LLM behavior in both monolingual settings ($g_1 = g_2$) and cross-lingual ($g_1 \neq g_2$) settings;
(2) comparing \textit{Few-shot} prompting using $g_1 \rightarrow g_2$ with \textit{Few-shot + FVs$^{g_1}$} to evaluate the ability of FVs to strengthen task representations in both monolingual and cross-lingual settings. 
Next, we describe the technique in detail, including FV extraction and injection in LLMs.

\subsection{FV Extraction}
As mentioned earlier, FV extraction uses a single language; therefore, we omit the parameter $g_1$ from all formulas for simplicity.
We adopt the FV formulation proposed by \citet{todd2024function}, which uses the \textit{causal mediation analysis} technique 
\cite{10.5555/2074022.2074073,NEURIPS2020_92650b2e,wang2023interpretability}--activation patching--to identify the attention heads responsible for encoding task information in few-shot prompts. \citet{todd2024function} showed that there exists a subset of attention heads that encodes task information from few-shot demonstrations, and that these heads are fixed and shared across tasks.
Therefore, we first extract this subset of FV heads using a simple extractive task (\texttt{Concept\_V\_Object\_5}\footnote{\texttt{Concept\_V\_Object\_5} is one of the 21 extractive-style tasks introduced in the original FV work \cite{todd2024function}.}). These heads are then used to compute the average activation projected onto the residual stream, which serves as the \textit{multi-label emotion detection FV} for subsequent experiments.

\paragraph{FV Head Extraction.} To construct a control condition, a pair consisting of an \textit{informative} prompt ($p$) and an \textit{uninformative} (known as \textit{corrupted}) prompt $\tilde{p}$ is created by randomly shuffling labels:
\begin{align}
p &= \left[ (x_{1}, y_1), \cdots, (x_{N}, y_{N}), x_{q} \right] \\ 
\tilde{p} &= \left[ (x_{1}, \tilde{y}_{1}), \cdots, (x_{N}, \tilde{y}_{N}), x_{q} \right]
\end{align}
Given a transformer model $f$, the mean task-conditioned activation for each attention head $a_{\ell j}$ is computed and cached at the last token position in input informative prompts:
\begin{equation}
\bar{a}_{\ell j} = \frac{1}{|\mathcal{P}|} \sum_{p \in \mathcal{P}} a_{\ell j}(p).
\end{equation}
where $a_{\ell j}(p)$ is the activation of layer $\ell$, head $j$ when the model processes the clean prompt ($p$), $\mathcal{P}$ is the set of clean prompts. The causal contribution of each attention head (causal indirect effect) is estimated by measuring how well an attention head can restore the model's ability to predict the target label when its activation is replaced with the average clean activation. The \textit{average indirect effect} (AIE) is computed over corrupted prompts, yielding a ranking score for each attention head.
\begin{equation} \small
\mathrm{AIE}_{\ell j}   = \frac{1}{|\tilde{\mathcal{P}}|} 
\sum_{\tilde{p} \in \tilde{\mathcal{P}}}   \big(
f(\tilde{p} \mid a_{\ell j} = \bar{a}_{\ell j})[y_{q}] 
- f(\tilde{p})[y_{q}]\big)
\end{equation}
The set of influential attention heads $\mathcal{A}$ (FV heads) is selected based on the top-$k$ highest AIE scores across all heads. Note that,  $\mathrm{AIE}_{\ell j}$ is computed based on the \texttt{Concept\_V\_Object\_5} task.

\paragraph{Multi-label Emotion Detection FV.} 
We construct the set of clean prompts for the multi-label emotion detection task, denoted by $\mathcal{P}^{e}$, to compute the average activation for this task. Given the set of FV heads $\mathcal{A}$ obtained in the previous step, the FV is constructed by aggregating the mean activations of the selected influential heads.
\begin{align}
\bar{a}^e_{\ell j} &= \frac{1}{|\mathcal{P}^e|} \sum_{p \in \mathcal{P}^e} a_{\ell j}(p) \\
v_k &= \sum_{{\ell j} \in \mathcal{A}} \bar{a}^{e}_{\ell j}.
\label{equation_k}
\end{align} 
where $k = |\mathcal{A}|$ is the hyperparameter used to select the FV heads in the previous step. Note that, for each language $g$, the set of clean prompts $\mathcal{P}^{e}$ is reconstructed and used to compute the corresponding activations and FVs (denoted $v_{kg}$ for language $g$).

\subsection{FV Injection}
To inject extracted FVs ($v_{kg}$) into the latent space of LLMs, we add them directly to the residual stream, following \citet{todd2024function}.
\begin{align}
h_\ell &= h_{\ell -1} + m_\ell + a_\ell + v_{kg}
\end{align}
where $h_\ell$ is the residual stream of the LLM at layer $\ell$, $m_\ell, a_\ell$, and $v_{kg}$ denote the projection of the MLP layer, attention head, and the extracted FV on the residual stream, respectively. Notably, the layer $\ell$ is selected to intervene also based on the corresponding $\mathrm{AIE}_{\ell j}$ scores computed from the previous step. 

\subsection{Prompt Construction \label{prompt_construction}}
To evaluate the task-steering effect of FVs, we consider two zero-shot prompting settings: a \textit{standard clean prompt} and a \textit{perturbed prompt}. 
The clean prompt contains only task instructions and candidate emotion labels, serving as a standard zero-shot baseline for evaluating the model's instruction-following capability.
The perturbed prompt additionally introduces auxiliary distractor labels (e.g., colors and nations) alongside the target emotion labels.
This controlled perturbation reduces the model's reliance on explicit prompt instructions and enables a more direct evaluation of whether FVs can recover the intended task behavior under misleading contextual cues.
Examples are shown in Tables~\ref{tab:few_shot},~\ref{tab:zero_shot} and~\ref{tab:standard_zero_shot} (Appendix~\ref{sec:ap_prompt}). 

\section{Experimental Setup}

\paragraph{Dataset}

To evaluate the multilingual generalization of FVs, we utilize the dataset released by the SemEval Task 11 \cite{muhammad-etal-2025-semeval}, which consists of sentence-level emotion annotations in 28 languages.
In our experiments, as summarized in Table~\ref{tab:dataset_statistics}, we select five typologically diverse languages: English (EN), German (DE), Chinese (ZH), Spanish (ES), and Russian (RU). Each instance is annotated with zero or more emotion categories, including joy, sadness, fear, anger, surprise, and disgust, forming a multilingual multi-label emotion classification task.

\begin{table}[htbp]
\centering
\small
\setlength{\tabcolsep}{6pt}
\caption{Statistics of the multilingual emotion classification datasets used in our experiments.}
\label{tab:dataset_statistics}
\begin{tabular}{lcccc}
\toprule
\textbf{Language} & \textbf{Train} & \textbf{Dev} & \textbf{Test} & \textbf{Total} \\
\midrule
Chinese (ZH) & 2,642 & 200 & 2,642 & 5,484 \\
German (DE) & 2,603 & 200 & 2,604 & 5,407 \\
English (EN) & 2,768 & 116 & 2,767 & 5,651 \\
Spanish (ES) & 1,996 & 184 & 1,695 & 3,875 \\
Russian (RU) & 2,679 & 199 & 1,000 & 3,878 \\
\bottomrule
\end{tabular}
\end{table}

\paragraph{Evaluation}
To ensure consistency with prior studies and enable fair comparisons, we follow the evaluation protocol of the SemEval Task 11 \cite{muhammad-etal-2025-semeval} and report the macro-averaged F1 score:
\begin{align}
\mathrm{F1}_{macro}
&=
\frac{1}{C}
\sum_{c=1}^{C}
\mathrm{F1}_c
\end{align}
where $C$ denotes the number of emotion classes and $\mathrm{F1}_c$ represents the F1 score of class $c$.

\paragraph{Experimental Environments}
We conduct experiments using two representative open-source LLMs: \texttt{Qwen3-8B} \cite{yang2025qwen3} and \texttt{Llama-3.1-8B-Instruct} \cite{grattafiori2024llama}. Unless otherwise specified, each experiment is repeated with \textit{five random seeds}, and all reported results are presented as mean$_{\pm\mathrm{std}}$ over the five runs. For the zero-shot setting, we evaluate only the first generated token under the greedy decoding strategy, whereas for the few-shot setting, the maximum number of new tokens is set to 10, since the model is expected to follow the demonstration format and generate a list of emotions.

\section{Results}

We evaluate FVs under monolingual and cross-lingual zero-shot settings with clean and perturbed prompts, as well as multilingual few-shot prompting with and without FV injection. As a baseline, we also measure zero-shot performance without FVs to assess whether FVs extracted from a source language can transfer task behavior to different target languages.

\subsection{Effectiveness of FVs -- Zero-shot Setting}

\paragraph{Standard Clean Prompt. \label{clean_prompt}}
We first evaluate FV injection under the \textit{standard clean prompt} (example in Table~\ref{tab:standard_zero_shot}, Appendix~\ref{sec:ap_prompt}). As shown in Table~\ref{tab:qwen_crosslingual_clean} and Table~\ref{tab:llama_crosslingual_clean} (Appendix~\ref{sec:zero_llama}), FV injection consistently improves performance over the standard clean zero-shot baseline across all evaluated languages and both LLMs. Since the task objective is explicitly specified in the instruction of the clean prompt, these improvements indicate that FVs provide additional task-relevant representations beyond standard prompt-based instruction following.

\begin{table}[t]
\centering
\setlength{\tabcolsep}{5pt}
\caption{
Macro-F1 scores (\%) of \texttt{Qwen3-8B} on the cross-lingual emotion classification task with \textbf{standard clean zero-shot} prompts. The first row reports the baseline without FVs \textit{(w/o FV)}, while the remaining rows use FVs extracted from random 5-shot demonstrations using the top-$k$ heads ($|\mathcal{A}| = 20$). Rows indicate the source languages for FV extraction ($g_1$), and columns denote the target evaluation languages ($g_2$).}
\label{tab:qwen_crosslingual_clean}
\resizebox{\linewidth}{!}{
\begin{tabular}{lccccc}
\toprule
\multirow{2}{*}{\textbf{FVs Src.}} & \multicolumn{5}{c}{\textbf{Target Language}} \\
\cmidrule(lr){2-6}
& \textbf{EN} & \textbf{DE} & \textbf{ZH} & \textbf{ES} & \textbf{RU} \\
\midrule
w/o FV & 36.4 & 16.9 & 18.4 & 41.4 & 52.5 \\
\midrule
\textit{(+FVs)} \\
EN &
\second{53.7$_{\pm 1.4}$} &
\best{44.1$_{\pm 0.6}$} &
\best{44.2$_{\pm 0.6}$} &
\second{62.4$_{\pm 1.0}$} &
\second{72.1$_{\pm 2.8}$} \\

DE &
\fourth{50.5$_{\pm 0.8}$} &
\fourth{41.9$_{\pm 0.9}$} &
\fourth{43.3$_{\pm 0.6}$} &
\fourth{60.3$_{\pm 1.4}$} &
\fourth{70.7$_{\pm 3.5}$} \\

ZH &
\third{51.5$_{\pm 1.1}$} &
\third{41.3$_{\pm 1.4}$} &
\third{43.5$_{\pm 0.2}$} &
\third{60.5$_{\pm 1.2}$} &
\third{71.2$_{\pm 1.2}$} \\

ES &
\best{54.4$_{\pm 1.1}$} &
\second{42.9$_{\pm 1.0}$} &
\second{43.9$_{\pm 0.6}$} &
\best{62.8$_{\pm 1.5}$} &
\third{72.4$_{\pm 2.2}$} \\

RU &
\third{52.1$_{\pm 1.7}$} &
\second{43.0$_{\pm 1.0}$} &
\third{43.6$_{\pm 1.3}$} &
\third{62.3$_{\pm 1.8}$} &
\best{74.5$_{\pm 1.2}$} \\
\bottomrule
\end{tabular}
}
\end{table}

\paragraph{Perturbed Prompt.}
To rigorously evaluate whether FVs can independently induce task behavior beyond explicit prompt instructions, we conduct experiments under the \textit{perturbed prompting} setting introduced in Section~\ref{prompt_construction} (example in Table~\ref{tab:zero_shot}, Appendix~\ref{sec:ap_prompt}). 
As shown in Table~\ref{tab:cross_lingual_zero_shot}, the perturbed zero-shot setting without FVs achieves low Macro-F1 scores across all languages, with performance close to random prediction in several cases. This indicates that the model struggles to infer the intended task behavior when explicit task cues are weakened. 
After injecting FVs, performance improves substantially across all evaluated languages. For example, the English target setting improves from 0.6 to 50.2, while the Chinese setting improves from 0 to 42.7. These results suggest that FVs capture task-relevant representations that effectively steer the model toward the target emotion classification behavior under highly perturbed prompting conditions.
Furthermore, the improvements remain consistent in cross-lingual transfer settings. Although using FVs derived from the same language as the target occasionally yields slightly stronger performance (e.g., Spanish (ES) and Russian (RU)), the differences are relatively small. Notably, in some cases (e.g., English (EN), Chinese (ZH) and German (DE)), cross-lingual transfer even outperforms same-language settings. These findings suggest that FVs encode transferable task-level representations rather than relying primarily on language-specific lexical patterns. 

Case studies in Table~\ref{tab:case_study} further illustrate this phenomenon. Without FVs, the perturbed prompts often lead to predictions from irrelevant label spaces, whereas applying cross-lingual FVs guides the model toward the correct emotion labels. Overall, these results demonstrate that FVs can transfer task behavior across languages without requiring target-language demonstrations during inference.
Additional cross-lingual zero-shot results for Llama are provided in Appendix~\ref{sec:zero_llama} (Table~\ref{tab:cross_lingual_zero_shot_llama}).

\begin{table}[t]
\centering
\setlength{\tabcolsep}{5pt}
\caption{
Macro-F1 scores (\%) of \texttt{Qwen3-8B} on the cross-lingual emotion classification task with \textbf{perturbed zero-shot} prompts. The FVs were extracted from random 5-shot demonstrations using the top-$k$ heads ($|\mathcal{A}| = 20$). 
}
\label{tab:cross_lingual_zero_shot}
\resizebox{\linewidth}{!}{
\begin{tabular}{lccccc}
\toprule
\multirow{2}{*}{\textbf{FVs Src.}} & \multicolumn{5}{c}{\textbf{Target Language}} \\
\cmidrule(lr){2-6}
  & EN & DE & ZH & ES & RU \\
\midrule
w/o FV
& 0.6 & 0.5 & 0.0 & 8.0 & 1.8 \\
\midrule
\textit{(+ FVs)}\\
EN &
\third{48.6 $_{\pm1.7}$} &
\second{38.2 $_{\pm0.9}$} &
\fourth{41.4 $_{\pm0.7}$} &
\third{51.8 $_{\pm2.6}$} &
\third{58.9 $_{\pm3.2}$} \\

DE &
\fourth{45.4 $_{\pm1.8}$} &
\fourth{35.6 $_{\pm1.2}$} &
\fourth{40.7 $_{\pm1.3}$} &
\fourth{49.4 $_{\pm1.8}$} &
\fourth{57.8 $_{\pm2.9}$} \\

ZH &
\fourth{46.3 $_{\pm1.7}$} &
\fourth{36.2 $_{\pm1.0}$} &
\second{42.2 $_{\pm0.7}$} &
\fourth{51.0 $_{\pm1.5}$} &
\fourth{58.8 $_{\pm3.2}$} \\

ES &
\best{50.2 $_{\pm0.9}$} &
\third{37.8 $_{\pm0.3}$} &
\second{42.2 $_{\pm0.5}$} &
\best{54.0 $_{\pm1.1}$} &
\second{60.3 $_{\pm1.4}$} \\

RU &
\second{49.4 $_{\pm1.1}$} &
\best{39.2 $_{\pm0.6}$} &
\best{42.7 $_{\pm0.9}$} &
\second{53.9 $_{\pm1.7}$} &
\best{64.6 $_{\pm2.7}$} \\
\bottomrule
\end{tabular}
}
\end{table}

\begin{table}[t]
\centering
\small
\caption{Case studies comparing perturbed zero-shot inference with and without cross-lingual FV injection.}
\label{tab:case_study}
\setlength{\tabcolsep}{4pt}
\renewcommand{\arraystretch}{1.15}
\begin{tabular}{p{0.95\linewidth}}
\toprule
\textit{(Example 1)} \\
\textbf{Sentence:}  \textit{``My heart dropped and I just replied `No.' ''} \\
\textbf{Gold Label:} fear \\
\textbf{Zero-shot Output (without FVs):} yellow \\
\textbf{Zero-shot Output (with FVs):} fear \\
\midrule
\textit{(Example 2)} \\
\textbf{Sentence:} \textit{``I slammed my fist against the door and yelled, Open up!''} \\
\textbf{Gold Label:} anger, fear \\
\textbf{Zero-shot Output (without FVs):} The man \\
\textbf{Zero-shot Output (with FVs):} anger \\
\bottomrule
\end{tabular}
\end{table}

\subsection{Effectiveness of FVs -- Few-shot Setting}

\begin{table}[t]
\centering
\setlength{\tabcolsep}{5pt}
\caption{
Macro-F1 scores (\%) of \texttt{Qwen3-8B} on the cross-lingual emotion classification task with \textbf{few-shot} prompts. The FVs were extracted from random 5-shot demonstrations in the source language using the top-$20$ heads. The denotation $^*$ indicates the results reported by previous works in SemEval 2025 Task 11: PAI \cite{ruan-etal-2025-pai} and JNLP \cite{xue-etal-2025-jnlp}. 
}

\resizebox{\linewidth}{!}{
    \begin{tabular}{p{0.16\linewidth}ccccc}
    \toprule
   \multirow{2}{*}{\textbf{FVs Src.}}  & \multicolumn{5}{c}{\textbf{Target Language}} \\
    \cmidrule(lr){2-6}
     & {EN} & {DE} & {ZH} & {ES} & {RU} \\
     \midrule
     \multicolumn{6}{c}{\cellcolor{Blue!6}{Previous works}}\\
    PAI$^*$  & 82.3 & 73.9 & 70.9 & 84.8  & 88.2 \\
    JNLP$^*$  & 80.4 & 69.9 & 68.1& 83.0 & 89.1 \\
    \midrule
     \multicolumn{6}{c}{\cellcolor{Blue!6}{This work}}\\
    w/o FV & 66.6 & 64.1 & 54.1 & 71.6 & 73.9 \\
    \textit{(+ FVs)}\\
    EN & \best{\textbf{67.6$_{\pm 0.3}$}} & \second{59.1$_{\pm 1.2}$} & 50.3$_{\pm 2.0}$ & \second{72.2$_{\pm 1.1}$} & 67.7$_{\pm 3.4}$ \\
    DE & \second{60.2$_{\pm 1.5}$} & \best{\textbf{64.6$_{\pm 0.5}$}} & \second{55.3$_{\pm 1.2}$} & \third{71.4$_{\pm 1.0}$} & \third{75.0$_{\pm 1.4}$} \\
    ZH & \fourth{57.3$_{\pm 1.5}$} & \fourth{44.3$_{\pm 4.2}$} & \fourth{53.5$_{\pm 1.2}$} & 62.5$_{\pm 1.6}$ & \fourth{73.7$_{\pm 1.4}$} \\
    ES & \third{59.8$_{\pm 0.9}$} & \third{55.7$_{\pm 0.9}$} & \third{54.7$_{\pm 0.7}$} & \best{\textbf{76.0$_{\pm 0.7}$}} & \best{\textbf{78.4$_{\pm 0.5}$}} \\
    RU & 52.6$_{\pm 1.3}$ & 43.2$_{\pm 2.0}$ & \best{\textbf{55.9$_{\pm 1.0}$}} & \fourth{63.4$_{\pm 1.2}$} & \second{76.2$_{\pm 1.5}$} \\
    \bottomrule
    \end{tabular}
}
\label{tab:fewshot_baseline}
\end{table}

\begin{table}[!htbp]
\centering
\small
\setlength{\tabcolsep}{5pt}
\caption{
Macro-F1 scores (\%) of \texttt{Qwen3-8B} with standard \textbf{5-shot} prompts without FV injection in \textbf{monolingual} and \textbf{cross-lingual} settings.
Rows indicate the source language (demonstration languages), while columns indicate the target evaluation languages.
}
    \begin{tabular}{p{0.2\linewidth}ccccc}
    \toprule
  \multirow{2}{*}{\textbf{Demo. Lang.}}  & \multicolumn{5}{c}{\textbf{Target Language}} \\
    \cmidrule(lr){2-6}
     & EN & DE & ZH & ES & RU \\
    \midrule
    EN &
    \best{66.6} &
    \fourth{45.2} &
    49.0 &
    \third{68.8} &
    65.9 \\
    
    DE &
    \second{61.6} &
    \best{64.1} &
    \second{53.7} &
    \second{70.5} &
    \second{74.0} \\
    
    ZH &
    \fourth{52.9} &
    \third{45.3} &
    \best{54.1} &
    \fourth{54.0} &
    \fourth{67.7} \\
    
    ES &
    \third{59.2} &
    \second{53.3} &
    \fourth{50.4} &
    \best{71.6} &
    \best{77.4} \\
    
    RU &
    51.8 &
    44.9 &
    \third{53.0} &
    55.9 &
    \third{73.9} \\
    \bottomrule
    \end{tabular}
\label{tab:all_fewshot}
\end{table}

Standard few-shot prompting provides strong performance by introducing explicit demonstrations during inference. We therefore investigate whether FV injection can further enhance few-shot ICL by providing additional latent task representations.
As shown in Table~\ref{tab:fewshot_baseline}, injecting FVs consistently improves performance across both monolingual and cross-lingual few-shot settings. 
Compared with standard few-shot prompting (Table~\ref{tab:all_fewshot}), FV-enhanced prompting achieves consistent gains across different language configurations. 
Interestingly, several cross-lingual FV settings even outperform the corresponding monolingual few-shot baselines. For example, transferring FVs extracted from Spanish to Russian achieves the best Russian performance, surpassing standard Russian few-shot prompting. Similar improvements are observed across multiple language pairs, suggesting that FV-induced representations are not restricted to the demonstration language.
These results demonstrate that FVs complement standard in-context learning by providing additional task-relevant representations. Although FVs do not completely replace the benefits of explicit demonstrations, they partially reproduce the task-steering effect of few-shot ICL while enabling cross-lingual transfer without requiring target-language demonstrations.

Compared with prior works \cite{ruan-etal-2025-pai, xue-etal-2025-jnlp}, which improves performance via ensembles of large fine-tuned LLMs such as \textit{Qwen 32B}, our method performs inference-time intervention on much smaller 8B LLMs, yet still achieves remarkable performance across languages.
\subsection{Analysis}
\subsubsection{Cross-lingual Alignment of FVs}
\begin{figure}[!htbp]
    \centering 
    \includegraphics[width=.89\linewidth, keepaspectratio, 
            trim={ 0 0 0 0 }, page=1, clip=true]{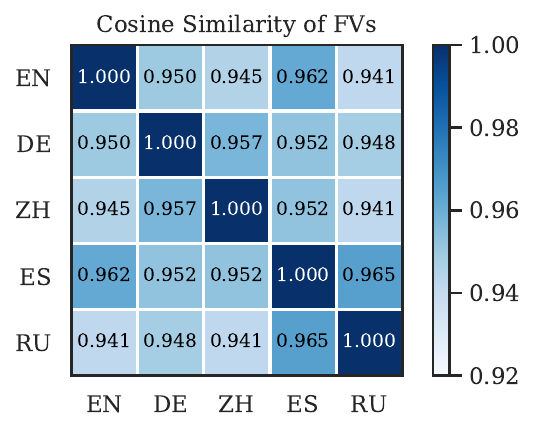}   
    \caption{Pairwise similarity scores of FV construction across different source languages ($v_{kg}$ for $g\in\{\mathrm{eng},\mathrm{deu},\mathrm{chn},\mathrm{esp},\mathrm{rus}\}$). Results are obtained with \texttt{Qwen3-8B} (5-shot, head = 20) and are reported as mean$_{\pm\mathrm{std}}$ over five random seeds.}
\label{fig:qwen_fv_similarity}
\end{figure}

To better understand why FVs extracted from one language can generalize to other languages, we analyze the similarity between FVs constructed from different source languages. Specifically, we measure the pairwise cosine similarity between FVs extracted from different source languages over five random seeds.
The results are summarized in Figure~\ref{fig:qwen_fv_similarity} and Figure~\ref{fig:llama_fv_similarity} (Appendix~\ref{sec:similar}). 
Across all language pairs, the cosine similarities are consistently high. For \texttt{Qwen3-8B}, the similarities range from 0.94 to 0.96, while for \texttt{Llama-3.1-8B} they range from 0.85 to 0.93, with relatively small standard deviations across random seeds.
These observations suggest that FVs extracted from different source languages occupy highly similar directions in the model representation space. This supports our hypothesis that cross-lingual transfer is enabled by a shared latent functional representation rather than language-specific vectors. In other words, although FVs are extracted from different languages, they capture largely overlapping functional directions, which provides a possible explanation for why FVs extracted from one language remain effective when injected into another language.








\subsubsection{Effect of FV Configuration on Robustness}
We conduct a series of ablation studies to investigate how different FV configurations affect intervention performance.

\paragraph{Number of Demonstrations for FV Constructions.} We first evaluate the robustness of FV injection under different \textit{numbers of demonstrations} (5-shot to 9-shot). As shown in Table~\ref{tab:qwen_crosslingual_clean_kshot}, FV injection consistently improves performance across all tested few-shot configurations, indicating that the resulting gains are not specific to any particular number of demonstrations. Moreover, the performance variation across different shot counts remains small, further confirming the stability of FV injection with respect to the number of demonstrations.

\begin{figure}[!htbp]
    \centering

    \begin{subfigure}[b]{0.93\linewidth}
        \centering
        \includegraphics[
            width=\linewidth,
            clip
        ]{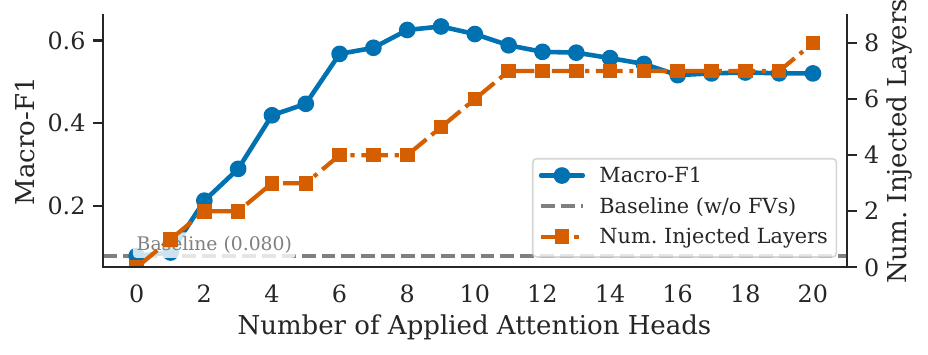}
        \caption{Monolingual (esp $\rightarrow$ esp)}
        \label{fig:fv_espesp}
    \end{subfigure}

    \vspace{0.2em}

    \begin{subfigure}[b]{0.93\linewidth}
        \centering
        \includegraphics[
            width=\linewidth,
            clip
        ]{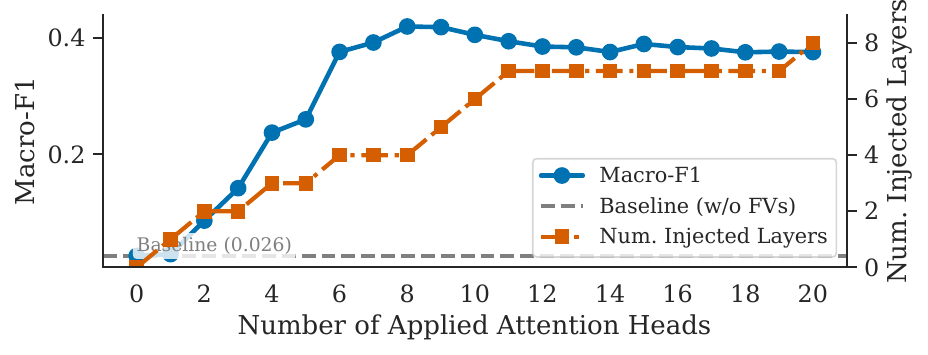}
        \caption{Cross-lingual (esp $\rightarrow$ deu)}
        \label{fig:fv_espdeu}
    \end{subfigure}

    \vspace{0.2em}

    \begin{subfigure}[b]{0.93\linewidth}
        \centering
        \includegraphics[
            width=\linewidth,
            clip
        ]{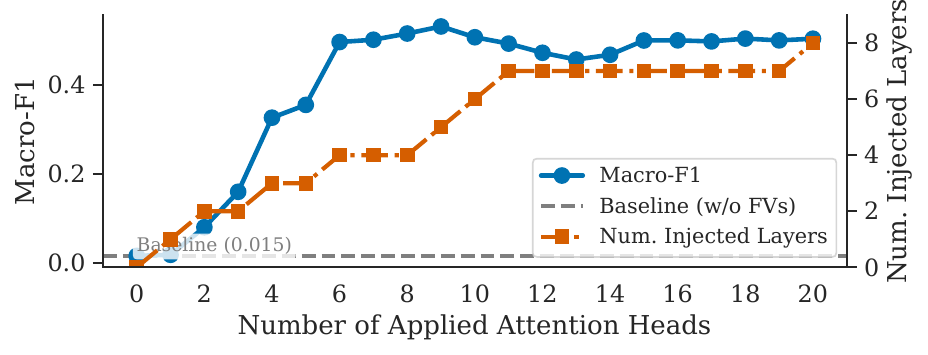}
        \caption{Cross-lingual (esp $\rightarrow$ eng)}
        \label{fig:fv_espeng}
    \end{subfigure}

    \caption{
   Effect of the top-$k$ attention heads used to construct FVs on performance under monolingual and cross-lingual settings with perturbed zero-shot prompts using \texttt{Qwen3-8B}.
    }

    \label{fig:fv_heads_all}
\end{figure}

\begin{table}[t]
\centering
\setlength{\tabcolsep}{5pt}
\caption{
Macro-F1 scores (\%) of \texttt{Qwen3-8B} on the cross-lingual emotion classification task with \textbf{standard clean zero-shot} prompts. The FVs were extracted from random demonstrations in the range from $5$-shot to $9$-shot in the source language using the top-$20$ heads.  
\label{tab:qwen_crosslingual_clean_kshot}
}

\resizebox{\linewidth}{!}{
\begin{tabular}{p{0.2\linewidth}ccccc}
\toprule
\multirow{2}{*}{\textbf{FVs Src. }} & \multicolumn{5}{c}{\textbf{Target Language}} \\

\cmidrule(lr){2-6}

& {EN} & {DE} & {ZH} & {ES} & {RU} \\
\midrule
w/o FV &
36.4 & 16.9 & 18.4 & 41.4 & 52.5 \\
\midrule
\textit{(+ FVs)}\\
EN &
\second{52.9$_{\pm 1.3}$} &
\best{43.7$_{\pm 0.6}$} &
\third{43.7$_{\pm 0.4}$} &
\fourth{61.6$_{\pm 1.1}$} &
\fourth{71.2$_{\pm 1.5}$} \\

DE &
\third{52.2$_{\pm 0.8}$} &
\third{43.3$_{\pm 0.6}$} &
\best{44.6$_{\pm 0.7}$} &
\second{62.2$_{\pm 0.9}$} &
\best{74.4$_{\pm 1.4}$} \\

ZH &
\fourth{50.3$_{\pm 0.3}$} &
\fourth{42.4$_{\pm 1.8}$} &
\second{44.3$_{\pm 0.6}$} &
\third{61.6$_{\pm 0.8}$} &
\second{73.3$_{\pm 1.0}$} \\

ES &
\best{55.0$_{\pm 0.7}$} &
\third{43.2$_{\pm 0.6}$} &
\third{43.8$_{\pm 0.7}$} &
\best{62.5$_{\pm 1.1}$} &
\third{71.4$_{\pm 1.1}$} \\

RU &
\third{51.9$_{\pm 0.9}$} &
\second{43.6$_{\pm 1.2}$} &
\third{43.4$_{\pm 0.8}$} &
\third{62.0$_{\pm 1.7}$} &
73.7$_{\pm 1.5}$ \\
\bottomrule
\end{tabular}}
\end{table}
\paragraph{Number of FV Heads.} We analyze the influence of the number of attention heads used for FV construction under both monolingual (Figure~\ref{fig:fv_espesp}) and cross-lingual settings (Figures~\ref{fig:fv_espdeu} and~\ref{fig:fv_espeng}) on \texttt{Qwen3-8B}.
The results consistently demonstrate that FV performance is strongly influenced by the number of selected attention heads. In general, incorporating more informative heads leads to stronger and more stable performance. However, after approximately six heads in \texttt{Qwen3-8B}, the performance gains begin to plateau, suggesting that the most effective steering signals are concentrated within a limited subset of attention heads, while additional heads contribute only marginal improvements. 
We further observe that the optimal head-selection pattern remains highly consistent across language pairs within the same LLM. Although the optimal number of heads varies across models, the overall trend is remarkably stable across multilingual transfer settings. Corresponding results for  \texttt{Llama-3.1-8B-Instruct} are provided in Appendix~\ref{sec:fv_head_llama} (Figure~\ref{fig:fv_heads_all_llama}). These observations suggest that the optimal FV configuration is largely model-specific but remains stable across source languages, allowing the same configuration to generalize across multilingual transfer settings.

\paragraph{Number of FV-injected layers.} We analyze how the number of FV-injected layers -- determined by the layers spanned by the selected top-$k$ function-vector heads-- relates to model performance (Figures~\ref{fig:fv_espesp},~\ref{fig:fv_espdeu} and \ref{fig:fv_espeng}). Prior work has shown that FVs applied to a single layer are often insufficient for semantically complex tasks such as sentiment analysis \cite{todd2024function}. Consistent with this observation, our results show that when the selected heads span only one or two layers, the resulting improvements are limited, whereas configurations in which the selected heads are distributed across multiple layers substantially enhance performance. 
These findings suggest that multi-layer FV injection facilitates the propagation of task-relevant representations, enabling more effective steering for complex classification tasks.

\paragraph{Number of FV Heads in Few-shot Setting.} We further investigate whether the above observations remain valid when FV injection is combined with few-shot prompting. Figure~\ref{FV_in_fewshot_qwen3_8b} (and Figure~\ref{FV_in_fewshot_qwen3_8b_apd} in Appendix~\ref{sec:few_shot_appendix}) presents the performance under different top-$k$ head selections.
Across both monolingual and cross-lingual settings, FV-enhanced few-shot prompting consistently outperforms the corresponding standard few-shot baseline across a broad range of top-$k$ configurations. These results indicate that FV-enhanced few-shot prompting remains effective across a broad range of top-$k$ selections, suggesting that its benefits do not rely on a narrowly tuned head configuration.

 \begin{figure*}[!htbp]
    \centering 
    \includegraphics[width=.99\linewidth, keepaspectratio, 
            trim={ 0 0 0 0 }, page=1, clip=true]{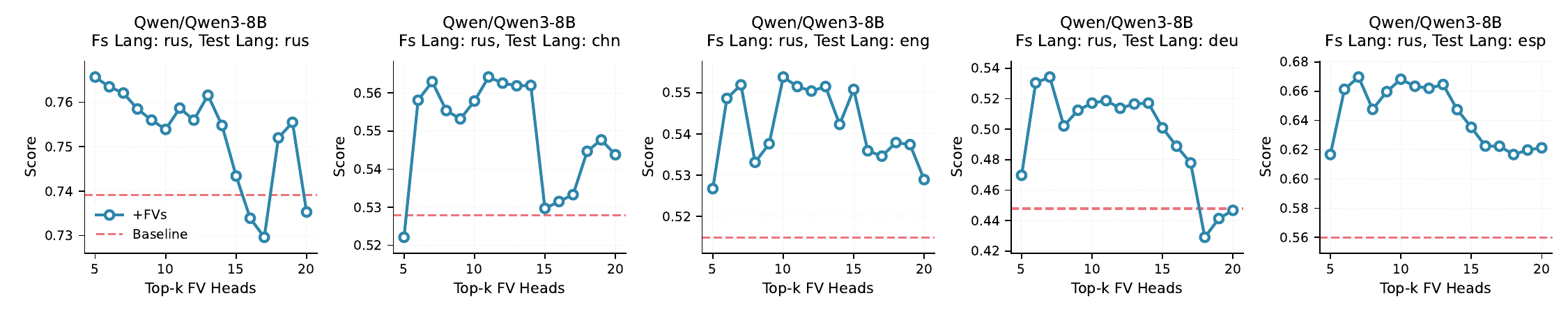} 
    \includegraphics[width=.99\linewidth, keepaspectratio, 
            trim={ 0 0 0 0 }, page=1, clip=true]{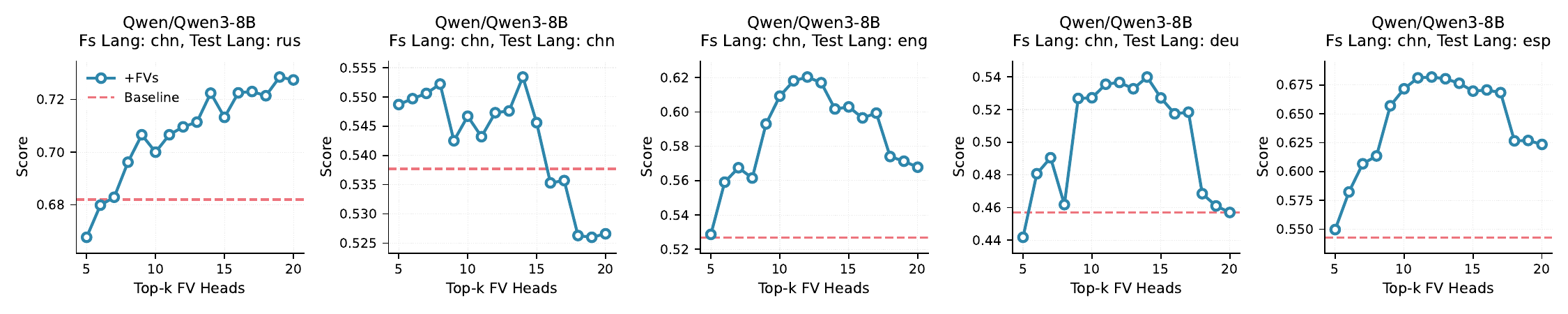} 
    \includegraphics[width=.99\linewidth, keepaspectratio, 
            trim={ 0 0 0 0 }, page=1, clip=true]{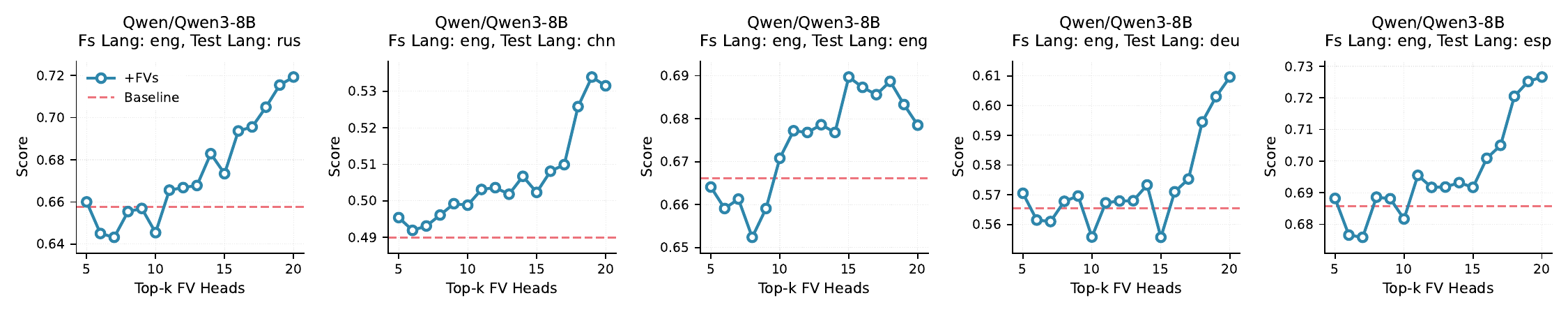} 
    \caption{
Performance of FV-enhanced few-shot prompting across different Top-$k$ head selections on \texttt{Qwen3-8B}. X-axis: number of selected FV heads ($k$); y-axis: classification performance score. The dashed line indicates the standard few-shot baseline without FV injection. ``Fs Lang'' ($g_1$): language used for FV extraction and few-shot demonstrations; ``Test Lang'' ($g_2$): language used in both demonstrations and query inputs during evaluation.
}
    \label{FV_in_fewshot_qwen3_8b}
\end{figure*}

\subsubsection{Error Analysis}

\begin{figure}[!h]
    \centering
    \includegraphics[
        width=\linewidth,
        trim={0cm 0.2cm 0cm 0cm},
        clip
    ]{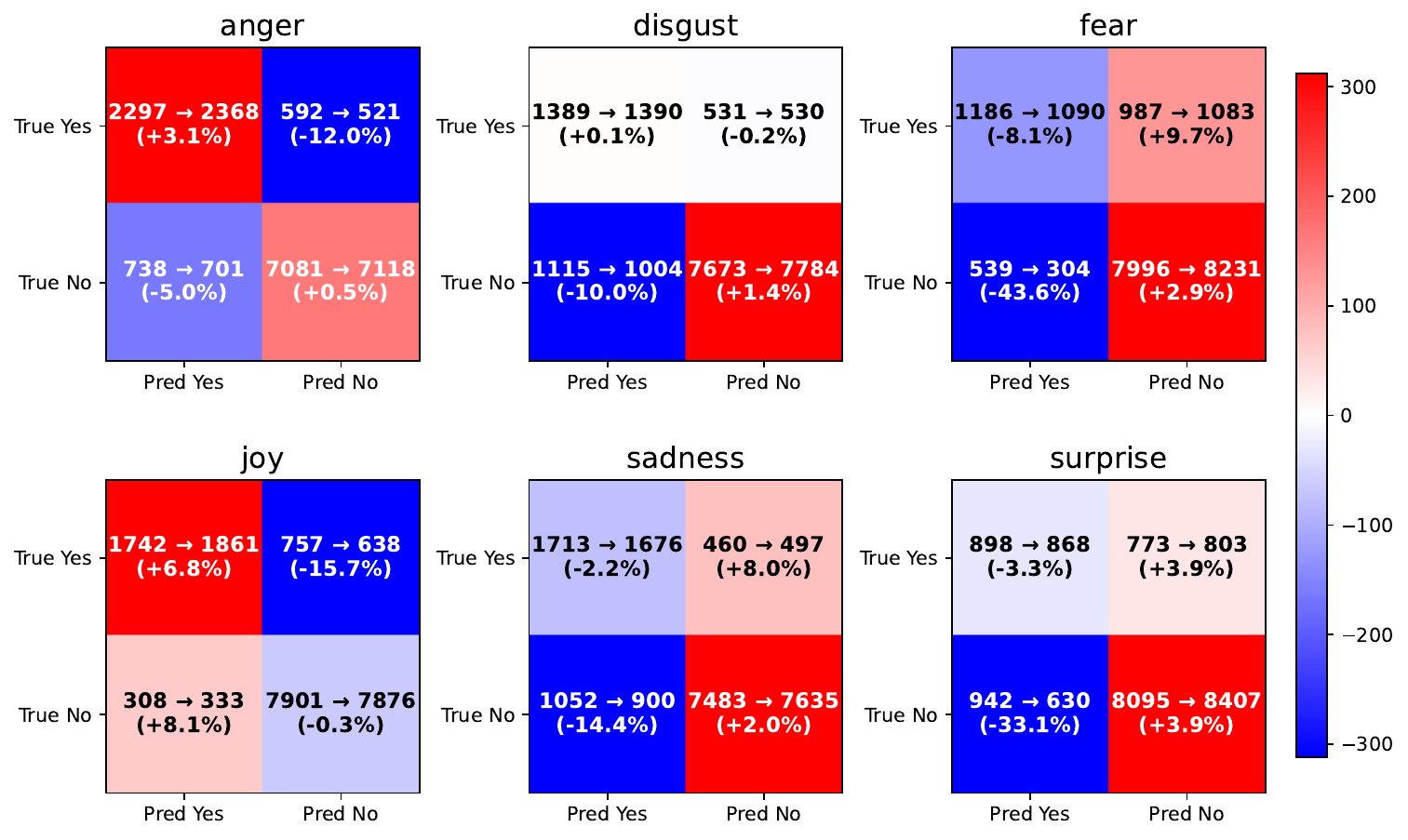}
    \caption{Confusion matrix across different emotion categories, where each emotion category combines results from all languages.}
    \label{fig:error_analysis}
\end{figure}

We analyze the impact of FV injection on prediction behavior through confusion matrix analysis across different emotion categories.
As illustrated in Figure~\ref{fig:error_analysis}, we aggregate the confusion matrix variations across five languages after applying FVs under the standard few-shot setting. Specifically, we analyze the changes in true positive (TP), true negative (TN), false positive (FP), and false negative (FN) to further investigate the precision--recall trade-off introduced by FVs.
Several important observations can be derived from the results. First, for relatively strong and explicitly expressed emotions such as \textit{anger} and \textit{disgust}, FVs lead to consistently improved classification behavior, where TP and TN increase while FN and FP decrease simultaneously. This indicates that FVs enhance both recognition sensitivity and decision reliability for strong emotional expressions.
Second, the substantial increase in TP together with the decrease in FN for \textit{joy} suggests that FVs enable the model to become more sensitive to \textit{joyful} emotional signals, although this also causes a small proportion of other emotional labels to be misclassified as \textit{joy}.
Moreover, for more subtle and easily confused emotions such as \textit{fear}, \textit{sadness}, and \textit{surprise}, the model becomes more careful and tends to avoid over-predicting these categories. This behavior is clearly reflected by the substantial decrease in FP, together with the increase in TN across these three emotions. These results suggest that FVs help the model make more cautious and reliable predictions for emotionally ambiguous categories.

Overall, the comparative confusion matrix analysis demonstrates that FVs consistently improve the model's prediction behavior across multiple emotion categories and languages, leading to more balanced and reliable emotion recognition performance.

\section{Conclusion}
In this paper, we investigated the cross-lingual transferability of FVs through multilingual multi-label emotion recognition, a semantically complex classification task. Our experimental results demonstrate that FVs exhibit strong cross-lingual generalization capabilities, enabling effective zero-shot task steering across languages without requiring in-context demonstrations during inference. Notably, cross-lingual FVs achieved substantial performance improvements, suggesting that FVs capture language-agnostic task-relevant representations rather than purely language-specific patterns. Furthermore, FVs can partially replicate the task-steering effects typically achieved through standard few-shot ICL, offering a more efficient alternative that circumvents the computational overhead associated with processing multiple demonstrations. These results establish FVs as a lightweight, transferable, and scalable mechanism for multilingual task adaptation, with significant implications for deploying large language models in cross-lingual scenarios where demonstration examples may be limited or unavailable.

\section*{Acknowledgments}
This work was partially supported by JSPS KAKENHI (No. 26K03016), and JST CREST (No. JPMJCR2563), JST CRONOS (No. JPMJCS24K7), AMED under Grant Number JP256f0137001, and JSPS KAKENHI Grant-in-Aid for Scientific Research (A), Grant Number JP25H00459.

\section*{Limitations}

Although we evaluate under a multilingual multi-label setting, our study primarily focuses on task-level cross-lingual transfer rather than label dependency modeling.
We do not explicitly investigate how FVs capture correlations among co-occurring emotion labels in multi-label prediction.
Our study focuses on multilingual emotion recognition as a classification (or multi-label classification) task, and it remains unclear whether the observed cross-lingual transferability of FVs generalizes to other multilingual reasoning or generation tasks (e.g., machine translation). In particular, generation tasks involve long output sequences, and over-steering by FVs throughout this process may corrupt the model's latent space, potentially degrading generation quality.




\bibliography{custom}

\appendix

\section{AI Usage Declaration }

AI tools were used for grammar checking and formatting of tables and figures, and polish writing. All technical content
and implementations were written by the authors.

\section{Prompting Content \label{sec:ap_prompt}}

\begin{table}[htbp]
    \centering
    \small
    \caption{English few-shot prompting template used for multilingual emotion classification. Blue text denotes the task instruction, while the red text indicates the target output generated by the LLM. Five demonstrations are used by default in the few-shot setting. \label{tab:few_shot}}
    \begin{tabular}{|p{0.98\columnwidth}|}
        \hline
        \textcolor{blue}{You are a multi-label emotion classification system. Each sentence contains zero or more emotions from the following list: anger, fear, joy, sadness, surprise, or none.}\\
        
        
        
        \#\#\# Emotions: anger, fear, joy, sadness, surprise, or none\\
        
        \#\#\# Input: For many months I learned to shower keeping my elbows close to prevent water from hosing off the ends and totally drenching the place.\\
        
        \#\#\#Answer: fear\\
        \\
        ... \texttt{\{\{demonstrations\}\}} ... \\
        \\
        
        
        \#\#\# Emotions: anger, fear, joy, sadness, surprise, or none\\
        
        \#\#\# Input: She commented on everything that went into my mouth.\\
        
        \#\#\# Answer: \color{red}{anger}\\
        \hline 

    \end{tabular}
\end{table}
\begin{table}[htbp]
    \centering
    \small
    \caption{Perturbed zero-shot prompting template for German ($\textit{g} = deu).$ \label{tab:zero_shot}}
    \begin{tabular}{|p{0.98\columnwidth}|}
        \hline 
        \#\#\# Colors: yellow, red, blue or none\\
        \#\#\# Nation: China, USA, Germany, England or none\\
        \#\#\# Emotions: anger, fear, joy, sadness, surprise, or none\\
        \#\#\# Input: Ich kenne keinen Staat, der Menschen, die unter seiner Herrschaft leben so unmenschlich behandelt wie.\\
        \#\#\#Answer: \\
        \hline 

    \end{tabular}
\end{table}

\begin{table}[htbp]
    \small
    \centering
    \caption{Standard zero-shot prompting template for German ($\textit{g} = deu).$ \label{tab:standard_zero_shot}}
    \begin{tabular}{|p{0.98\columnwidth}|}
    \hline
        \textcolor{blue}{You are a multi-label emotion classification system. Each sentence contains zero or more emotions from the following list: anger, fear, joy, sadness, surprise, or none.}\\
        \#\#\# Emotions: anger, fear, joy, sadness, surprise, or none\\
        \#\#\# Input: Ich kenne keinen Staat, der Menschen, die unter seiner Herrschaft leben so unmenschlich behandelt wie.\\
        \#\#\#Answer: \\
        \hline 

    \end{tabular}
\end{table}

\FloatBarrier


\section{{Cross-lingual Zero-shot Emotion Classification with Llama} \label{sec:zero_llama}}

\begin{table}[!htbp]
\centering
\setlength{\tabcolsep}{5pt}
\caption{
Macro-F1 scores (\%) of \texttt{Llama-3.1-8B-Instruct} on the cross-lingual emotion classification task with \textbf{perturbed zero-shot} prompts. The FVs were extracted from random 5-shot demonstrations using the top-$k$ heads ($|\mathcal{A}| = 5$).
\label{tab:cross_lingual_zero_shot_llama}
}

\resizebox{\linewidth}{!}{
\begin{tabular}{lccccc}
\toprule
\multirow{2}{*}{\textbf{FVs Src.}} & \multicolumn{5}{c}{\textbf{Target Language }} \\
\cmidrule(lr){2-6}
  & EN & DE & ZH & ES & RU \\
\midrule
w/o FV
& 14.1 & 12.9 & 8.4 & 29.1 & 35.5 \\
\midrule
\textit{(+ FVs)}\\
EN &
\fourth{40.9$_{\pm 4.2}$} &
\fourth{30.4$_{\pm 4.5}$} &
\fourth{38.7$_{\pm 2.0}$} &
\fourth{49.1$_{\pm 2.3}$} &
\fourth{46.7$_{\pm 5.6}$} \\

DE &
\third{42.8$_{\pm 6.3}$} &
\third{34.4$_{\pm 4.3}$} &
\third{40.0$_{\pm 2.0}$} &
\third{49.7$_{\pm 4.1}$} &
\third{51.4$_{\pm 4.9}$} \\

ZH &
\third{44.3$_{\pm 4.4}$} &
\third{34.5$_{\pm 2.4}$} &
\second{40.4$_{\pm 1.6}$} &
\third{49.2$_{\pm 2.0}$} &
\third{49.2$_{\pm 3.6}$} \\

ES &
\second{46.0$_{\pm 2.6}$} &
\second{37.1$_{\pm 1.5}$} &
\third{40.2$_{\pm 0.6}$} &
\second{53.4$_{\pm 1.7}$} &
\second{56.4$_{\pm 1.5}$} \\

RU &
\best{47.6$_{\pm 2.6}$} &
\best{38.6$_{\pm 1.7}$} &
\best{42.3$_{\pm 1.1}$} &
\best{55.6$_{\pm 1.7}$} &
\best{59.0$_{\pm 2.1}$} \\
\bottomrule
\end{tabular}
}
\end{table}

\begin{table}[!htbp]
\centering
\caption{
Macro-F1 scores (\%) of \texttt{Llama-3.1-8B-Instruct} on the cross-lingual emotion classification task with \textbf{standard clean zero-shot} prompts. The FVs were extracted from random 5-shot demonstrations using the top-$k$ heads ($|\mathcal{A}| = 5$).
\label{tab:llama_crosslingual_clean}
 }
\resizebox{\linewidth}{!}{
\begin{tabular}{lccccc}
\toprule
\multirow{2}{*}{\textbf{FVs Src.}} & \multicolumn{5}{c}{\textbf{Target Language }} \\
\cmidrule(lr){2-6}
  & EN & DE & ZH & ES & RU \\
\midrule
 w/o FV & 36.1 & 18.9 & 12.7 & 31.0 & 21.9 \\
\midrule
\textit{(+ FVs)}\\
EN &
\fourth{39.6$_{\pm 2.7}$} &
\fourth{29.1$_{\pm 2.1}$} &
\best{35.3$_{\pm 1.2}$} &
\fourth{44.4$_{\pm 2.7}$} &
\fourth{42.4$_{\pm 5.9}$} \\

DE &
\third{41.0$_{\pm 1.6}$} &
\second{29.5$_{\pm 1.7}$} &
\best{35.3$_{\pm 1.4}$} &
\second{45.7$_{\pm 1.5}$} &
\third{47.7$_{\pm 5.6}$} \\

ZH &
\second{41.3$_{\pm 1.4}$} &
\third{29.4$_{\pm 0.5}$} &
\third{33.3$_{\pm 2.9}$} &
\fourth{44.2$_{\pm 0.3}$} &
\fourth{45.4$_{\pm 2.8}$} \\

ES &
\second{42.3$_{\pm 0.6}$} &
\second{29.5$_{\pm 1.6}$} &
\fourth{32.7$_{\pm 1.1}$} &
\second{45.3$_{\pm 1.7}$} &
\second{50.8$_{\pm 1.0}$} \\

RU &
\best{43.0$_{\pm 1.2}$} &
\best{30.4$_{\pm 2.5}$} &
\fourth{32.8$_{\pm 3.1}$} &
\best{47.1$_{\pm 3.1}$} &
\best{53.3$_{\pm 3.9}$} \\

\bottomrule
\end{tabular}
}
\end{table}

\FloatBarrier

\section{Effect of Top-\textit{k} Attention Heads in Llama-3.1-8B-Instruct \label{sec:fv_head_llama}}

\begin{figure}[!t]
    \centering

    \begin{subfigure}[b]{0.99\linewidth}
        \centering
        \includegraphics[
            width=\linewidth,
            trim={0 0 0 0},
            clip
        ]{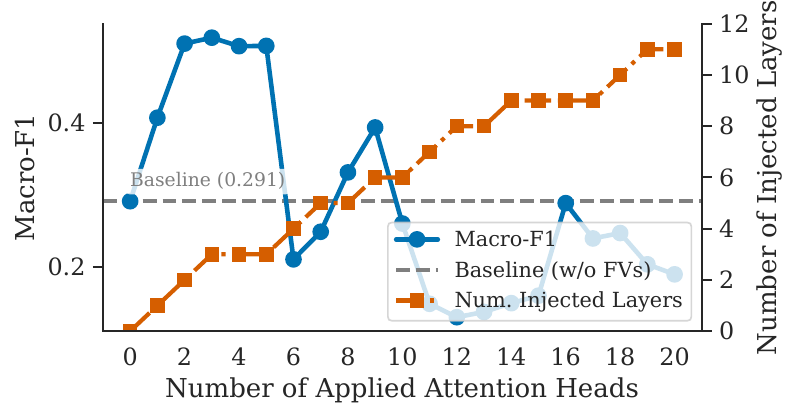} 
        \caption{Monolingual (esp $\rightarrow$ esp)}
        \label{fig:fv_espesp_llama}
    \end{subfigure}

    \vspace{0.2em}

    \begin{subfigure}[b]{0.99\linewidth}
        \centering
        \includegraphics[
            width=\linewidth,
            trim={0 0 0 0 },
            clip
        ]{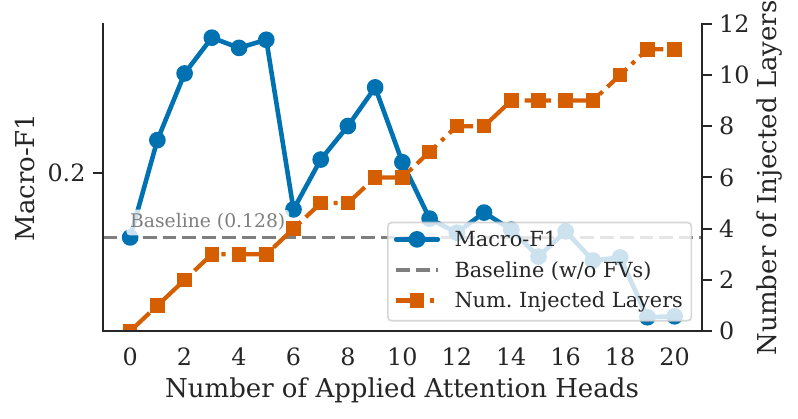} 
        \caption{Cross-lingual (esp $\rightarrow$ deu)}
        \label{fig:fv_espdeu_llama}
    \end{subfigure}

    \vspace{0.2em}

    \begin{subfigure}[b]{0.99\linewidth}
        \centering
        \includegraphics[
            width=\linewidth,
            trim={0 0 0 0},
            clip
        ]{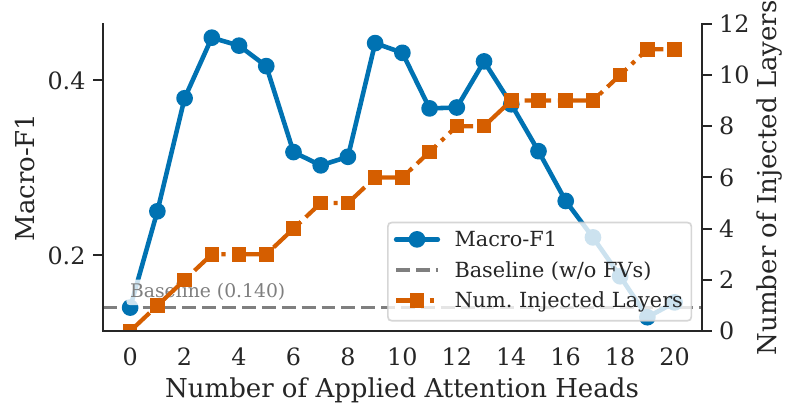} 
        \caption{Cross-lingual (esp $\rightarrow$ eng)}
        \label{fig:fv_espeng_llama}
    \end{subfigure}

    \caption{
    Effect of top-$k$ attention heads used for constructing FVs under monolingual and cross-lingual settings in \texttt{Llama-3.1-8B-Instruct}.
    }

    \label{fig:fv_heads_all_llama}
\end{figure}

\section{Additional Cross-lingual Few-shot Results \label{sec:few_shot_appendix}}

 \begin{figure*}[!htbp]
    \centering 
    \includegraphics[width=.99\linewidth, keepaspectratio, 
            trim={ 0 0 0 0 }, page=1, clip=true]{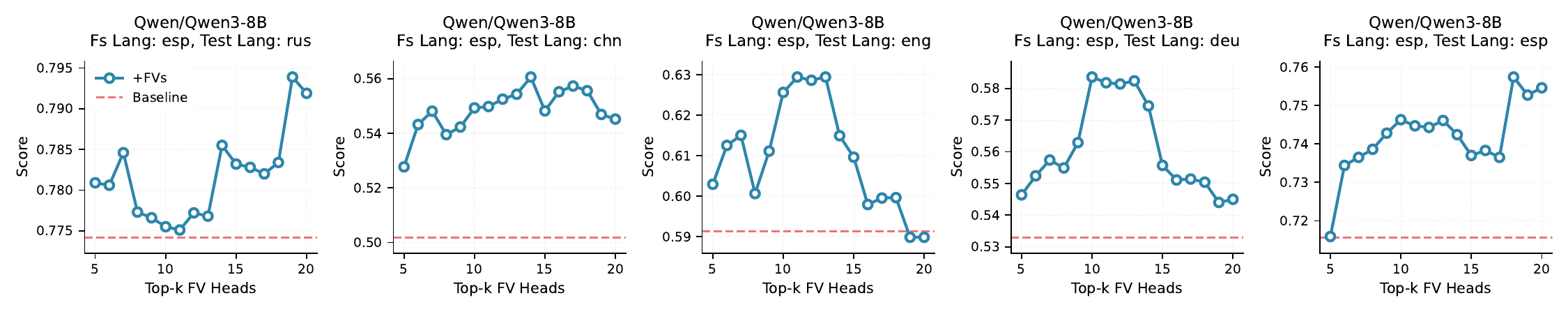} 
    \includegraphics[width=.99\linewidth, keepaspectratio, 
            trim={ 0 0 0 0 }, page=1, clip=true]{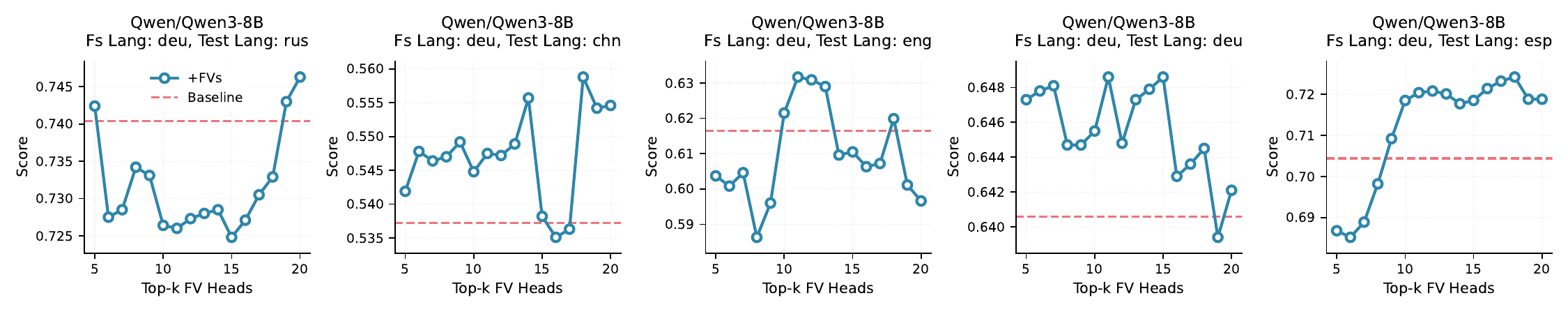} 
    \caption{Performance of FV-enhanced few-shot prompting across different Top-$k$ FV head selections on Qwen3-8B.
The x-axis denotes the number of selected FV heads ($k$), while the y-axis represents the classification performance score.
The dashed horizontal line indicates the corresponding standard few-shot prompting baseline without FV injection.
``Fs Lang'' ($g_1$) denotes the language used for FV extraction and few-shot demonstrations, and ``Test Lang'' ($g_2$) denotes the language used in both the demonstrations and query inputs during evaluation.}
    \label{FV_in_fewshot_qwen3_8b_apd}
\end{figure*}

\FloatBarrier

\section{Mechanism Analysis\label{sec:similar}}

 \begin{figure}[!htbp]
    \centering 
    \includegraphics[width=.99\linewidth, keepaspectratio, 
            trim={ 0 0 0 0 }, page=1, clip=true]{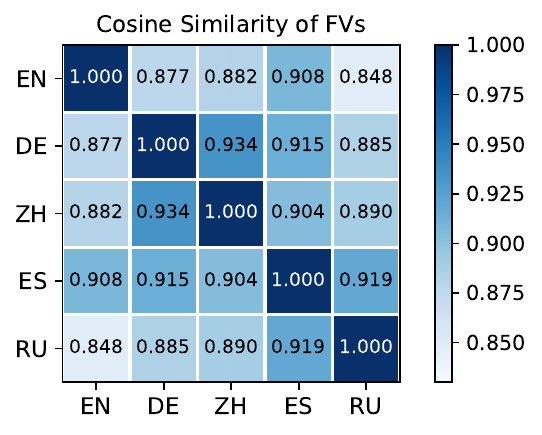}   
    \caption{Pairwise similarity scores of FV construction across different source languages ($v_{kg}$ for $g \in \{\mathrm{eng},\mathrm{deu},\mathrm{chn},\mathrm{esp},\mathrm{rus}\}$). Results are obtained with \texttt{Llama-3.1-8B-Instruct} (5-shot, head = 5).}
\label{fig:llama_fv_similarity}
\end{figure}










\FloatBarrier

\end{document}